\documentclass{article}

 \pdfoutput=1

\usepackage{arxiv}
\usepackage{algpseudocode}
\usepackage{tabularx}
\usepackage{algorithm}
\usepackage[caption=false]{subfig}
\usepackage{verbatim}
\usepackage[utf8]{inputenc} % allow utf-8 input
\usepackage[T1]{fontenc}    % use 8-bit T1 fonts
\usepackage{hyperref}       % hyperlinks
\usepackage{url}            % simple URL typesetting
\usepackage{booktabs}       % professional-quality tables
\usepackage{amsmath,amsfonts,amssymb}
\usepackage{graphicx}
\usepackage[numbers]{natbib}
\usepackage{doi}

\title{Automatic Segmentation of Left Ventricle in Cardiac Magnetic Resonance Images}

\author{
  Garvit Chhabra \\
  Department of Electrical and Electronics Engineering
  \\Manipal Institute of Technology, Manipal Academy of Higher Education
  \\ Manipal, India - 576104 \\
	\And
  J. H. Gagan \\
  Department of Electrical and Electronics Engineering
  \\Manipal Institute of Technology, Manipal Academy of Higher Education
  \\ Manipal, India - 576104 \\
  \And
  J. R. Harish Kumar \\
  Department of Electrical and Electronics Engineering,
  \\Manipal Institute of Technology, Manipal Academy of Higher Education
  \\ Manipal, India - 576104 }
  	
\date{}

\renewcommand{\shorttitle}{\textit Segmentation of Left Ventricle}

\hypersetup{
pdftitle={Automatic Segmentation of Left Ventricle in Cardiac Magnetic Resonance Images},
}
\begin{document}
\maketitle

\begin{abstract}
Segmentation of the left ventricle in cardiac magnetic resonance imaging MRI scans enables cardiologists to calculate the volume of the left ventricle and subsequently its ejection fraction. The ejection fraction is a measurement that expresses the percentage of blood leaving the heart with each contraction. Cardiologists often use ejection fraction to determine one's cardiac function. We propose multiscale template matching technique for detection and an elliptical active disc for automated segmentation of the left ventricle in MR images. The elliptical active disc optimizes the local energy function with respect to its five free parameters which define the disc. Gradient descent is used to minimize the energy function along with Green's theorem to optimize the computation expenses. We report validations on 320 scans containing 5,273 annotated slices which are publicly available through the Multi-Centre, Multi-Vendor, and Multi-Disease Cardiac Segmentation (M\&Ms) Challenge. We achieved successful localization of the left ventricle in 89.63\% of the cases and a Dice coefficient of 0.873 on diastole slices and 0.770 on systole slices. The proposed technique is based on traditional image processing techniques with a performance on par with the deep learning techniques.\end{abstract}


\section{Introduction}
\label{sec:introduction}
In a single period of the cardiac cycle of the heart, two phases occur - relaxation, also known as diastole, and contraction which is also known as systole. During diastole, blood flows to the heart and during systole, the blood flows out of the two ventricles (left and right) of the heart to the lungs and body. Regardless of the force of contraction, the heart can never pump out all the blood present in its ventricles \cite{hrt3}. The ejection fraction (EF) is the percentage of blood that is pumped out by the heart in each cycle. Although the EF can be calculated for both the ventricular chambers, in most cases it refers to the left ventricle (LV) since it is the one responsible for pumping out oxygenated blood to the entire body. An EF of less than 40\% may be evidence of heart failure or some form of cardiomyopathy whereas an EF of more than 70\% indicates the stiffness or abnormal thickening of the heart muscle which is known as hypertrophic cardiomyopathy \cite{hrt}. 

The EF is obtained by calculating the volume of blood in the LV at two different time frames during the cardiac cycle. These two quantities - the end of systole volume (ESV) and the end of diastole volume (EDV) are used to calculate the EF as follows:
\begin{comment}
\begin{equation}
\text{SV=EDV-ESV}
\end{equation} 
\end{comment}
\begin{equation}
\text{EF}=\frac{\text{EDV \, -- \, ESV}}{\text{EDV}} \times 100.
\end{equation}
\begin{figure*}[t]
\centering
\includegraphics[width=0.95\textwidth]{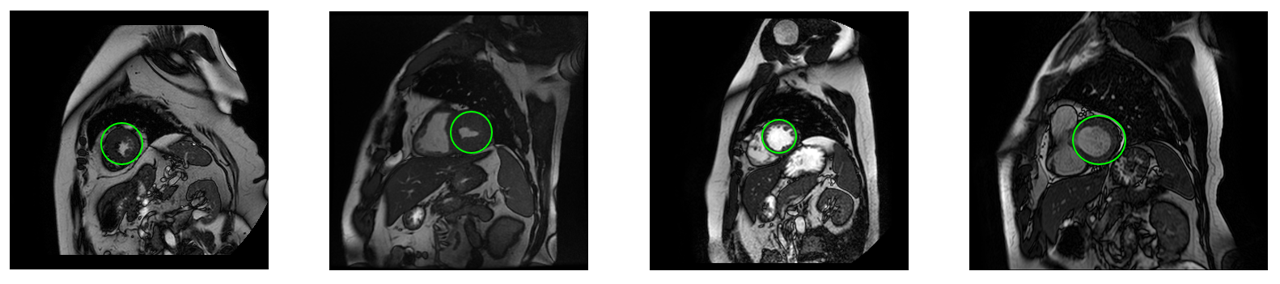}
\caption{Variability in cardiac MRI scans taken at different phases (time) and points on the short axis (z-axis).}
\label{lv}
\end{figure*}
The volume of the LV can be assessed using various imaging techniques like echocardiography, cardiac catheterization, cardiac computerized tomography (CT), cardiac magnetic resonance imaging (MRI), and cardiac nuclear medicine scan \cite{hrt2}. Cardiac MRI is generally considered to be the gold standard for volume estimation and three-dimensional modeling of the heart \cite{Mri}. To calculate the EF cardiologist segment the LV in MRI scans.
When done manually the cardiologists have to delineate the LV in multiple slices taken along its short axis, which is perpendicular to its long axis. These slices are then interpolated to obtain the volume of the LV. When done manually computation of EF is a tedious and time-consuming process that is also susceptible to inter-observer variations. Automating the process will allow for faster diagnosis and may result in more accurate results \cite{hrt4}. 

\subsection{Prior Work}
\label{subsec:lr}
In this section, we will discuss some of the notable methods that have been proposed for the segmentation of the LV. The segmentation of the LV in cardiac MR images can be broken down into two steps; locating the heart in MR scans and segmenting the LV. Localizing the heart reduces the computation required by allowing the segmentation algorithm to process only the localized region rather than the entire MR image. The proposed methodologies to detect the ROI can be classified into two: Time-based techniques and object detection techniques \cite{petijean}. In time-based techniques, the algorithm utilizes the temporal nature of the heart, as it is the only organ in the scan whose shape transforms with time. The work presented by Pednekar et al. \cite{pednekar} can be categorized as a time-based technique. In their technique, a difference image is computed from two images at two different points in the cardiac cycle. Hough transform is then applied to detect a circular region from the difference image. Cocosco et al. \cite{cocoso} computed the standard deviation of image intensity at each voxel (3D analogous of a pixel) and projected it along the z-axis. The result is binarized and dilated several times and then the convex hull is computed to obtain the ROI. In the previously discussed time-based methods, no prior information was needed, whereas in object detection techniques strong prior information is a key requirement. These methods learn features to train classifiers and detect where the heart is present in a particular image \cite{hog,sift}. 

The second step is the segmentation of the LV which is achieved by etching the inner wall (endocardium) and outer wall (epicardium) of the said ventricle. Delineating the endocardium is comparatively easier as it surrounds the blood pool (light) which provides a good contrast with the myocardium (dark) \cite{petijean}. Several methods have been proposed over the past few decades to segment the LV. Gupta et al. \cite{gupta} proposed a method using dynamic programming (DP), by assigning lower costs to object boundaries in a cost matrix. Lorenzo-Valdés et al. \cite{atlas} used non-rigid registration to deform a volumetric atlas. Kaus et al. \cite{KAUS} used a fully automated three-dimensional deformable model to segment the myocardium in MR images. Mitchell et al. \cite{mitchell} used a hybrid combination of active appearance models (AAMs) and active shape models (ASMs) to drive the fitting process for segmentation of left and right ventricles. Commercial software packages are also available for the task, such as MR analytical software system (MASS) and Argus (Siemens Medical Systems, Germany) \cite{vander}.

In the field of deep learning, convolutional neural networks (CNN) such as U-Net and its variants have been extensively used for the segmentation of the LV. Liao et al. \cite{dl} used a cascade classifier with CNN to detect the ROI and then used hypercolumns fully convolutional neural network (HFCN) to segment and estimate the volume of the LV. Deep learning techniques are capable of achieving high accuracy for medical image segmentation. However, they do tend to fail when tested on scans belonging to a domain different from the one they were trained on (e.g., scans from different clinical centers, scanners, etc.).  The participants of the Multi-Centre, Multi-Vendor, and Multi-Disease Cardiac Segmentation (M\&Ms) Challenge developed various strategies to counter this issue. Full et al. \cite{nnu} used the nnU-Net framework with extensive data augmentation and batch normalization to achieve great results with their model across different domains. Ma \cite{jun} utilized histogram matching to reduce the variance in cross-domain scans with Z-score (mean subtraction and division by standard deviation) to normalize the images. The processed images were then used to train a 3D U-Net model which is subsequently used for segmentation. Liu et al. \cite{liu} proposed two data augmentation methods i.e., resolution augmentation to rescale images to different resolutions and factor-based augmentation to decompose scans into two lateral representations: anatomical and modal, these representations are then used to generate new images. The resulting dataset is used to train the SDNet model used for segmentation. Corral Acero et al. \cite{unet1} also adopted a 2-step approach to segmentation i.e., heart detection and then segmentation. Additionally, domain adversarial training of neural networks (DANN) and domain unlearning (DU) were used to counter the problem of image heterogeneity. Li et al. \cite{unet2} augmented the domain using a random style transfer technique to dampen the impact of cross-domain scans. Zhang et al. \cite{zhang} used histogram matching to augment the domain, in addition to this, they also used unlabelled slices for label propagation to fine-tune the results. Scannell et al. \cite{scannell} trained a domain invariant 2D U-Net using domain adversarial learning. A domain discriminator was also trained using labeled and unlabeled data. Carscadden et al. \cite{carscadden} proposed an approach using U-Net++ and ResNet101 architectures. They applied a preprocessing step that maintained only the largest connected component in 3D space amongst the predictions made by the network. Saber et al. \cite{saber} proposed a U-Net model enhanced with dilated inception block and multi-gate block. This allowed them to input the image to the network in varying resolutions with different levels of detail. Khader et al. \cite{khader} proposed to resample the scans and set the voxel depth for all scans to the median voxel depth of the training set. The resulting augmented data was used to train a 3D U-Net model. Parreño et al. \cite{parreno} used a Resnet-34 model to classify the image and appropriately adapt it to a known domain using iterative backpropagation. The resulting adapted images were then segmented using U-Net. Li et al. \cite{drunet} utilized Cycle Generative Adversarial Network (CycleGAN) for one-to-one image translations to a target domain, with Dilated Residual U-Net (DRUNet) to segment the translated images. Kong \& Shadden \cite{kong} proposed a method using frequency domain augmentation and CycleGAN for data augmentation. An architecture based on attention U-Net was used which allowed them to repress the irrelevant features and focus only on the cardiac structures. Huang et al. \cite{huang} proposed a model to render style invariant images to remove the domain variations present in images from different vendors. A pretrained VGG-16 served as the backbone for style transfer while segmentation was achieved using 2D U-Net.

\subsection{Our Contribution}
\label{subsec:cont}
Our proposed method utilizes an elliptical active disc (EAD) for segmenting the LV in short-axis MRI scans. The formulation of the active disc is motivated by the work presented by Thevenaz et al. \cite{thev}, Pediredla et al. \cite{snakes}, and Kumar et al. \cite{ead}. The active disc is capable of segmenting scans from diverse domains if there is a distinguishable contrast between the endocardium and myocardium. Automatic initialization of the disc is achieved by using multiscale template matching technique. The aforementioned technique enables us to localize the LV in scans taken at different resolutions. The segmentation of LV is a precursor for the estimation of volume of the heart at diastole and systole and hence for the computation of EF.
\section{Methodology}
\label{sec:methodology}
We have adopted the two-step approach towards the LV segmentation problem as discussed in Section \ref{subsec:lr}. The heart is localized using multiscale template matching technique -- an extended version of template matching which allows us to detect the heart in MRI scans with different resolutions using a single template image \cite{mstm}. The LV is then segmented using an EAD which contains a pair of concentric discs that dynamically deform to minimize a given energy function. Since the EF can be calculated by segmenting the endocardium alone our approach will only focus on its segmentation. 
\subsection{Multiscale Template Matching}
\begin{figure*}[!t] %Multiscale image
\centering
\subfloat[]{\includegraphics[width=0.1\textwidth]{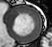}}\\
\subfloat[]{\includegraphics[width=0.95\textwidth]{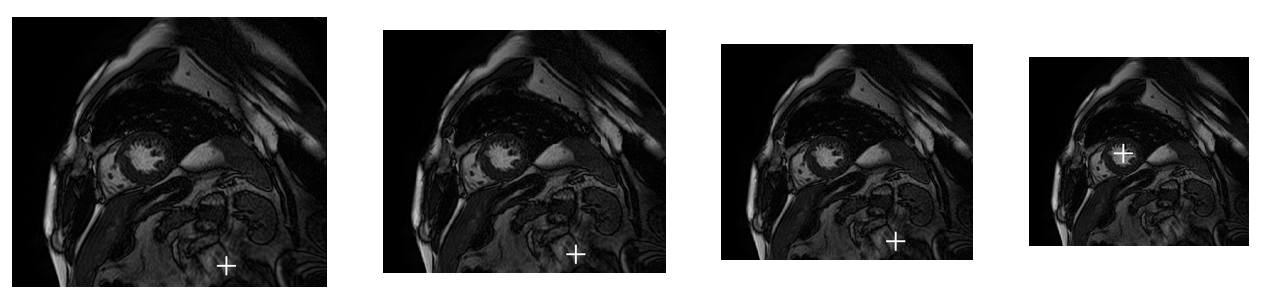}}
\caption{(a) Template image cropped from $288 \times 288$px slice. (b) multiscale template matching finding max similarity on different scaled versions (1x, 0.9x, 0.8x, 0.7x) of a $512 \times 440$px slice using the template image in (a).}
\label{ms}
\end{figure*}
A template matching algorithm traverses the template over the search image, and at each position it calculates a value to estimate the degree of similarity between the template and the region on the image spanned by the template. We have used the multiscale normalized template matching technique to detect the LV. It measures the similarity between the template and search image as a function of displacement relative to each other at multiple resolutions \cite{ncc2}. It is calculated using the following formula:
\begin{equation}
\hspace{-2mm}
\zeta(x,y)=\frac{\sum_{x_{t},y_{t}}(T(x_{t},y_{t}).I(x+x_{t},y+y_{t}))}{\sqrt{ {\sum_{x_{t},y_{t}}(T(x_{t},y_{t}))^2} .{\sum_{x_{t},y_{t}}(I(x+x_{t},y+y_{t}))^2}}},
\end{equation}
where $\zeta$ is the normalized cross-correlation coefficient with a value in the range $0\le \zeta \le 1$. $I(x+x_{t},y+y_{t})$ and $T(x_{t},y_{t})$ represent the search and template image respectively, with $x_t$ and $y_t$ being the relative displacement. 
\algnewcommand{\LineComment}[1]{\State \(\triangleright\) #1}
\begin{algorithm}[!t]
\caption{Multiscale Template Matching (MTM)}
\label{alg1}
\begin{algorithmic}
\State $scale \leftarrow 1$
\State $max \leftarrow 0$
\LineComment{Store original image's height and width}
\State $H_{org} \leftarrow$ getHeight(Image)  
\State $W_{org} \leftarrow$ getWidth(Image)
\While {$scale>0$}
\State $H_{temp} \leftarrow H_{org} \times scale$
\State $W_{temp} \leftarrow  W_{org} \times scale$
\State ImageScaled $\leftarrow$ resize(Image, $H_{temp}, W_{temp}$)
\LineComment{Apply normalized template matching on scaled image to find coordinates with maximum correlation coefficient}
\State $\zeta(x, y) \leftarrow$ MTM(ImageScaled, Template) 
\If{$\zeta(x, y)>max$} 
\State $max \leftarrow \zeta(x, y)$
\State $x_m \leftarrow x$
\State $y_m \leftarrow y$
\State $s_m \leftarrow scale$
\EndIf
\State $scale \leftarrow scale - 0.1$
\EndWhile
\State $X \leftarrow x_m/s_m $
\State $Y \leftarrow y_m/s_m $
\end{algorithmic}
\end{algorithm}
Multiscale template matching is an extension of template matching which allows matching in cases where the search and the image from which the template is sourced is not of the same resolution. Thus, multiscale template matching makes it possible to use a single template image to match cardiac MRI scans taken at different resolutions. In this technique, we match the template image with different scaled versions of the search image and retain the maximum correlation and its coordinates in a temporary variable. The scaling factor is then used to obtain the coordinates pertaining to the image with original dimensions. Algorithm \ref{alg1} details the working of the multiscale template matching algorithm. Fig. \ref{ms}. visualizes the working of the algorithm as it finds and stores the coordinates and scaling factor at which the maximum similarity is obtained.
\subsection{Elliptical Active Disc Design and Optimization}
\begin{figure}[!t]
\centering
\includegraphics[width=0.6\textwidth]{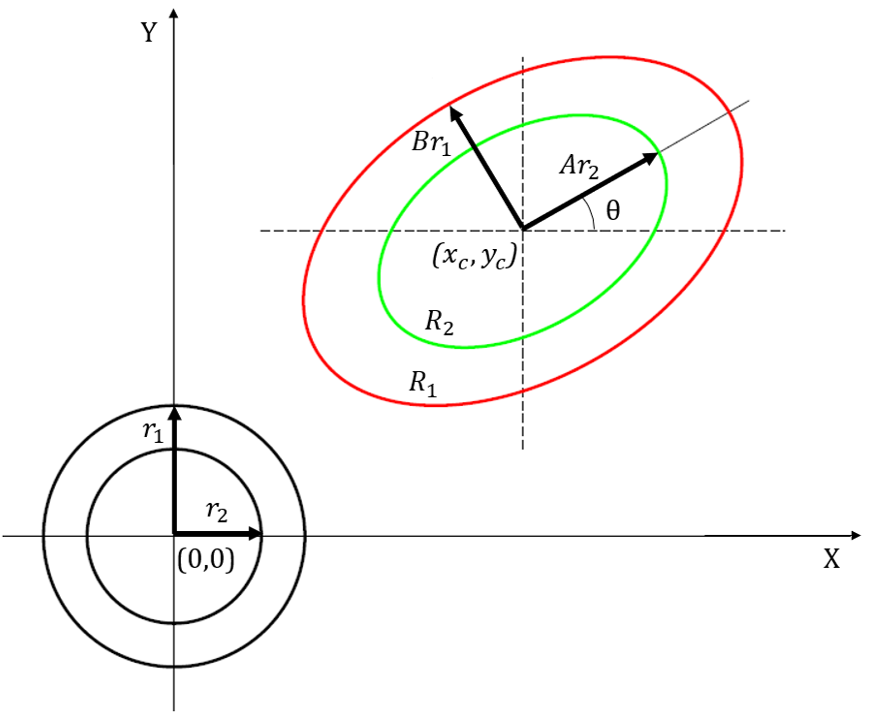}
\caption{The active disc template and its EAD.}
\label{disc}
\end{figure}

Once the LV is detected, an EAD will be used to segment and delineate it.
The disc template contains a pair of concentric circles. It is then subjected to an affine transformation that allows for horizontal and vertical scaling, rotation, and translation to make it an active disc. The active disc template with concentric circles is parameterized as below:
\begin{equation}
\begin{pmatrix} 
x_{i}(t) \\
y_{i}(t)
\end{pmatrix}
=
\begin{pmatrix} 
r_{i}\cos t\\
r_{i}\sin t
\end{pmatrix},
\label{ad}
\end{equation}
 where $\forall t \in (0,2\pi ]$, and $i$=1, 2, are used to represent the outer and inner circle whose radii are set equal to a ratio of  $1:1/\sqrt{2}$, which sets the area covered by the inner circle to be equal to the area covered by the annulus. The affine transformation of the disc template is given by,
\begin{equation}
\label{param}
\begin{pmatrix} 
X_{i} \\
Y_{i}
\end{pmatrix}
=
\begin{pmatrix} 
A\cos \theta & B\sin \theta \\
-A\sin \theta & B\cos \theta
\end{pmatrix}
\begin{pmatrix} 
x_{i}\\
y_{i}
\end{pmatrix}
+
\begin{pmatrix} 
x_{c}\\
y_{c}
\end{pmatrix},
\end{equation}
\noindent where $A$ and $B$ are the scaling parameters representing the semi-major axis and semi-minor axis of the ellipse, $\theta$ is the angle of rotation, and $x_c$ and $y_c$ are the translatory parameters representing the center coordinates of the concentric discs. An EAD exhibits dynamic behavior and actively works at maximizing a contrast function between the outer and inner circles. Fig. \ref{disc}. provides a visual representation of the disc template and its affine transformed version resulting in EAD from (\ref{ad}) and (\ref{param}), respectively. The normalized contrast function for an image $f(X, Y)$ with $R_1$ and $R_2$ being the regions enclosed by outer and inner discs, respectively, is given by  
\begin{equation}
E = \frac{1}{AB} \left ( \iint_{R_{1} \setminus R_{2}} f(X,Y) \,\mathrm{d}X\,\mathrm{d}Y  - \iint_{R_{2}} f(X,Y) \,\mathrm{d}X\,\mathrm{d}Y \right ),\nonumber
\end{equation}
\begin{equation}
E = \frac{1}{AB} \bigg ( \underbrace{\iint_{R_{1}} f(X,Y) \,\mathrm{d}X\,\mathrm{d}Y}_{E_{1}}  - ~~ 2\underbrace{\iint_{R_{2}} f(X,Y) \,\mathrm{d}X\,\mathrm{d}Y}_{E_{2}} \bigg ),\nonumber
\label{eq6_a} 
\end{equation}
\begin{equation}
E = \frac{1}{AB} (E_1 - 2E_2).
\label{eq6} 
\end{equation}
\noindent The optimal fit of EAD on the ROI is obtained by using the gradient descent algorithm which requires the computation of partial derivatives of $E$ with respect to the five free parameters. Using Green’s theorem the surface integral can be represented as line integral allowing for faster computation:
\begin{equation}
\label{green}
 E_2=\oint_{R_2} f^{X}\mathrm{d}Y = -\oint_{R_2} f^{Y}\mathrm{d}X.
\end{equation}
The partial derivative of $E_2$ can be calculated using the following:
\begin{equation}
\label{derive}
\frac{\partial E_{2}}{\partial A}=\frac{\partial E_{2}}{\partial X}\frac{\partial X}{\partial A}+\frac{\partial E_{2}}{\partial Y}\frac{\partial Y}{\partial A}.
\end{equation}
Substitute (\ref{green}) in (\ref{derive}):
\begin{equation}
\frac{\partial E_{2}}{\partial A}=\oint_{R_2}\frac{\partial f^{X}}{\partial X}\frac{\partial X}{\partial A}\mathrm{d}Y-\oint_{R_2}\frac{\partial f^{Y}}{\partial Y}\frac{\partial Y}{\partial A}\mathrm{d}X.
\label{eq9}
\end{equation}
From (\ref{eq9}), we get 
\begin{equation}
\frac{\partial E_{2}}{\partial A}=\oint_{R_2}f(X,Y)~ B~ x ~\mathrm{d}y.
\label{eq9_1}
\end{equation}
\noindent Substituting for $\mathrm{d}y$ and $x$ in  (\ref{eq9_1}), we get
\begin{equation}
\label{eq10}
\frac{\partial E_{2}}{\partial A}=\frac{B}{2} \left ( \int_{t=0}^{2\pi}f(X_{2},Y_{2}) \, \cos^2(t)\, \mathrm{d}t \right ).
\end{equation}
Similarly, we can obtain the partial derivative of $E_1$ with respect to A:
\begin{equation}
\label{eq11}
\frac{\partial E_{1}}{\partial A}=B \left (\int_{t=0}^{2\pi}f(X_{1},Y_{1}) \, \cos ^2(t) \,\mathrm{d}t \right ).
\end{equation}
Differentiating  (\ref{eq6}) with respect to A:
\begin{equation}
\label{eq12}
\frac{\partial E}{\partial A}=\frac{1}{AB} \left (\frac{\partial E_{1}}{\partial A} -2\frac{\partial E_{2}}{\partial A}\right )-\frac{1}{A^2B}(E_{1}-2E_{2}).
\end{equation}
Substituting (\ref{eq10}) and  (\ref{eq11}) in (\ref{eq12}),
\begin{equation}
\frac{\partial E}{\partial A}=\frac{1}{A} \left (\int_{t=0}^{2\pi}(f(X_{1},Y_{1})-f(X_{2},Y_{2})) \, \cos ^2(t) \,\mathrm{d}t -E \right ).\nonumber
\end{equation}
Similarly, we find the partial derivative of $E$ with respect to $B$ as,
\begin{equation}
\frac{\partial E}{\partial B}=\frac{1}{B} \left (\int_{t=0}^{2\pi}(f(X_{1},Y_{1})-f(X_{2},Y_{2})) \, \sin ^2(t) \, \mathrm{d}t -E \right ).\nonumber
\end{equation}
We also calculate the partial derivative of $E$ with respect to the angle of rotation and center coordinates:
\begin{equation}
\hspace{-6mm}
\frac{\partial E}{\partial \theta}=\frac{1}{2AB} \left (\int_{t=0}^{2\pi}(f(X_{1},Y_{1})-f(X_{2},Y_{2}))(B^2-A^2) \, \sin (2t) \,\mathrm{d}t \right ),\nonumber
\end{equation}
%\vspace{-6mm}
\begin{equation}
\frac{\partial E}{\partial x_{c}}=\frac{1}{AB} \left (\int_{t=0}^{2\pi}(\sqrt2 f(X_{1},Y_{1})-2f(X_{2},Y_{2}))\, \cos (t) \, \mathrm{d}t \right ),\nonumber
\end{equation}
\begin{equation}
\frac{\partial E}{\partial y_{c}}=\frac{1}{AB} \left (\int_{t=0}^{2\pi}(\sqrt2 f(X_{1},Y_{1})-2f(X_{2},Y_{2}))\, \sin (t) \,\mathrm{d}t \right ).\nonumber
\end{equation}
\section{Results}
An ImageJ \cite{IJ} plugin was developed to implement the proposed method. The database used for validation is from the M\&Ms Challenge which was organized as part of the Statistical Atlases and Computational Modelling of the Heart (STACOM) Workshop held in conjunction with the MICCAI 2020 Conference \cite{dataset}. The database consists of scans obtained from 6 different hospitals using scanners from 4 different vendors namely Siemens, Canon, GE, and Phillips. We have tested our technique across the entire database. Our method was tested on 320 scans containing 98,810 slices which were accessible through the M\&Ms challenge's online resource. Segmentation results were evaluated only on the 5,273 annotated slices. Annotations were only available for the slices at the end of diastole and end of systole. The annotated slices were used to compare the segmentation results obtained from the proposed technique. The segmented slices at the end of diastole and systole were interpolated to compute the volume and subsequently the EF. A mean absolute error of 12.03\% for the EF was observed on the scans tested. The parameters used to initialize the EAD remained constant for all the scans i.e., the parameters of the disc were not fine-tuned on a per scan basis. It can be inferred from Table \ref{tab:multi} that even though multiscale template matching is more accurate it's still prone to false localizations in cardiac MR scans. These false localizations lead to failed segmentation results rendering zero true positives, hence negatively impacting the segmentation performance of the proposed algorithm. Table \ref{tab:semi} provides a comparison between two implementations of our method one utilizing multiscale template matching and the other using manual inputs from a user to initialize the position of the LV (e.g., click-based initialization, input centre coordinates, etc.). This allows us to analyze the segmentation performance of the EAD by itself unaffected by the failures of multiscale template matching. A Dice coefficient of 0.873 on diastole slices and 0.770 on systole slices were obtained using the elliptical active disk technique. The Dice coefficient for all the 5,273 annotated slices was calculated to be 0.826.
\begin{table}[!t]
\centering
\renewcommand{\arraystretch}{1.1}
\caption{Comparison between multiscale and template matching}
\label{tab:multi}
 \begin{tabularx}{0.9\textwidth}{  >{ \centering \arraybackslash} X >{\hsize=0.6\hsize \centering \arraybackslash} X }
%\newcolumntype{Y}{>{\centering\arraybackslash}X}
    \hline \hline
    	Technique & Accuracy \\ \hline \hline
   	Template Matching  & 83.695\% \\ 		
        Multiscale Template Matching   & 89.130\% \\ \hline \hline
%Mean Absolute Error
  \end{tabularx}
\end{table}
\begin{table}[!t]
\centering
\caption{Comparison between semi-automatic and automatic segmentation}
\label{tab:semi}
\renewcommand{\arraystretch}{1.1}
 \begin{tabularx}{0.9\textwidth}{ >{ \centering \arraybackslash} X >{ \centering \arraybackslash} X >{\centering \arraybackslash} X }
%\newcolumntype{Y}{>{\centering\arraybackslash}X}
    \hline \hline
    Metric &Semi-Auto & Auto \\ \hline \hline
   	Jaccard Index & 0.718 &0.651 \\ 		
    Dice Coefficient & 0.826& 0.750 \\
	Sensitivity & 0.881& 0.801 \\ 
	Specificity &0.998& 0.997 \\ 
	Accuracy & 0.996& 0.992\\ 
	Precision & 0.803&0.765\\ \hline \hline
%mean absolute error \\
  \end{tabularx}
\end{table}
\begin{table}[!t]
\centering
\begin{minipage}{0.9\textwidth}
\begin{center}
\caption{Comparison with deep learning techniques from M\&Ms challenge}
\renewcommand{\arraystretch}{1.1}
 \begin{tabularx}{1\textwidth}{ X >{ \centering \arraybackslash} X >{\centering \arraybackslash} X }
%\newcolumntype{Y}{>{\centering\arraybackslash}X}
    \hline \hline
    Technique & DC-ED\footnote{Dice coefficient of end of diastole slices } & DC-ES\footnote{Dice coefficient of end of systole slices } \\ \hline
   \hline			
    nnU-Net \cite{nnu} & 0.939& 0.886\\ 
	U-Net \cite{unet1}&0.927& 0.877 \\ 
	DRUNET \cite{drunet}&0.922& 0.857\\ 
	SDNet \cite{liu}&0.889&0.835\\
	U-Net \cite{huang}&0.896&0.772\\ 
	U-Net \cite{unet2}&0.797& 0.716 \\
	Proposed EAD &0.873 & 0.770 \\ \hline \hline
  \end{tabularx}
\label{tab:compare}
\end{center}
\end{minipage}
\end{table}
Table \ref{tab:compare} provides a comparison between our EAD technique and some of the deep learning-based techniques that took part in the M\&Ms challenge. The results in Table \ref{tab:compare} were obtained from Campello et al. \cite{dataset}. The techniques from the competition were tested on 160 scans whose specifics are not known to us. Hence we tested our technique on all the 320 scans available to us through the M\&Ms website and no scans were used for training. In Fig. \ref{fig_res}.  the segmented results of two scans - G7Q2W0 and N7P3T8 are shown. These scans were captured using MRI scanners from two different vendors that are Phillips and Canon, respectively. 
\begin{figure*}[!t] %RESULT
\centering
\subfloat[At the end of diastole in scan G7Q2W0 (z=1 to z=9).]{\includegraphics[width=0.9\textwidth]{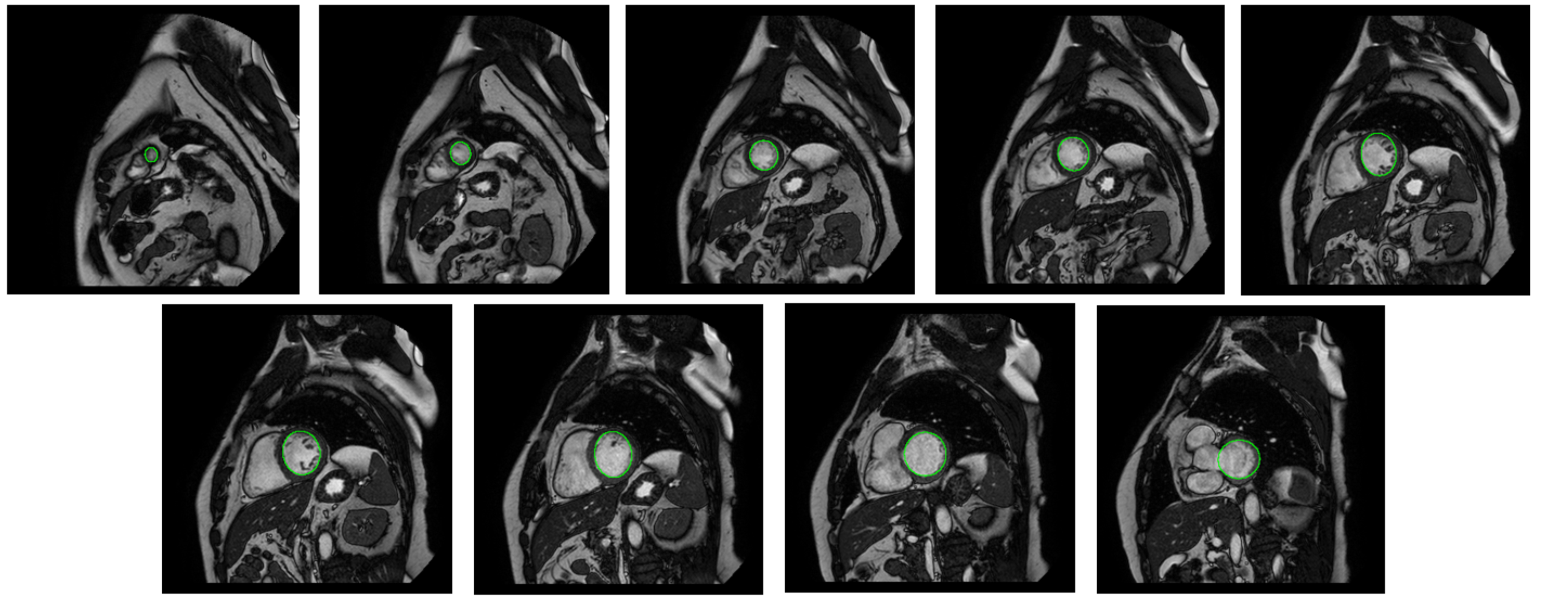}
\label{Diastole}}\\
\subfloat[At the end of systole in scan G7Q2W0 (z=1 to z=8).]{\includegraphics[width=0.9\textwidth]{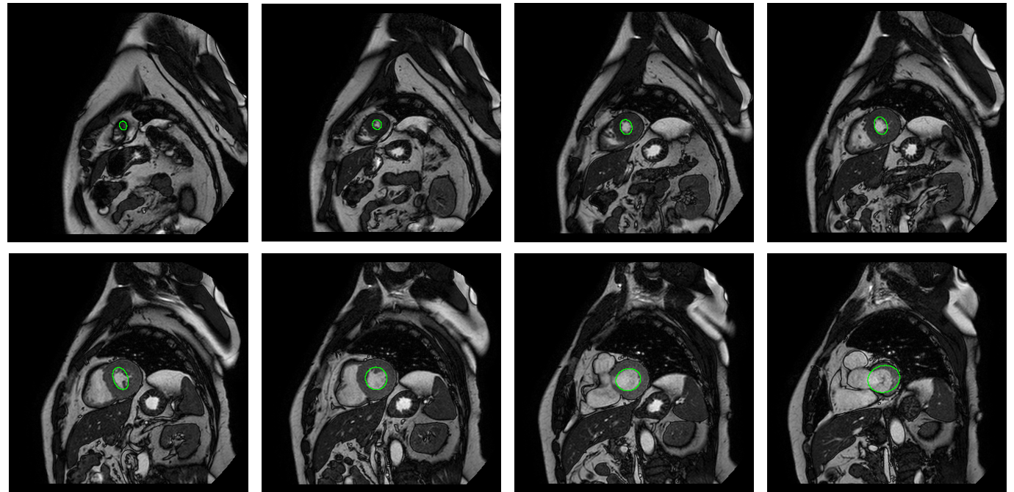}
\label{Systole}}\\
\subfloat[At the end of diastole in scan N7P3T8 (z=2 to z=11).]{\includegraphics[width=0.9\textwidth]{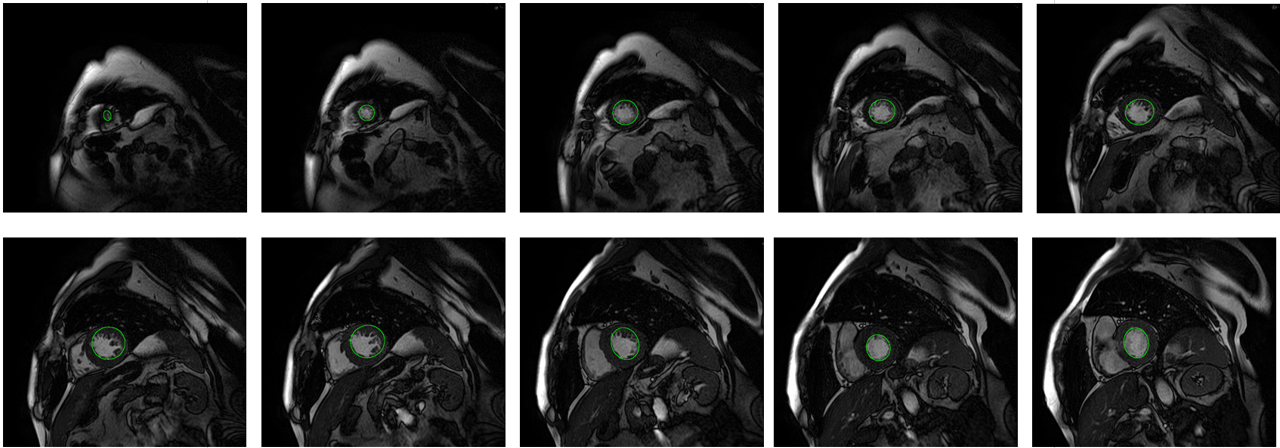}
\label{Diastole2}}\\
\caption{Segmented results of two different scans captured from two different scanners.}
\vspace{10mm}
\end{figure*}
\begin{figure*}[!t]
\centering
\ContinuedFloat
\subfloat[At the end of systole in scan N7P3T8 (z=3 to z=10).]{\includegraphics[width=0.85\textwidth]{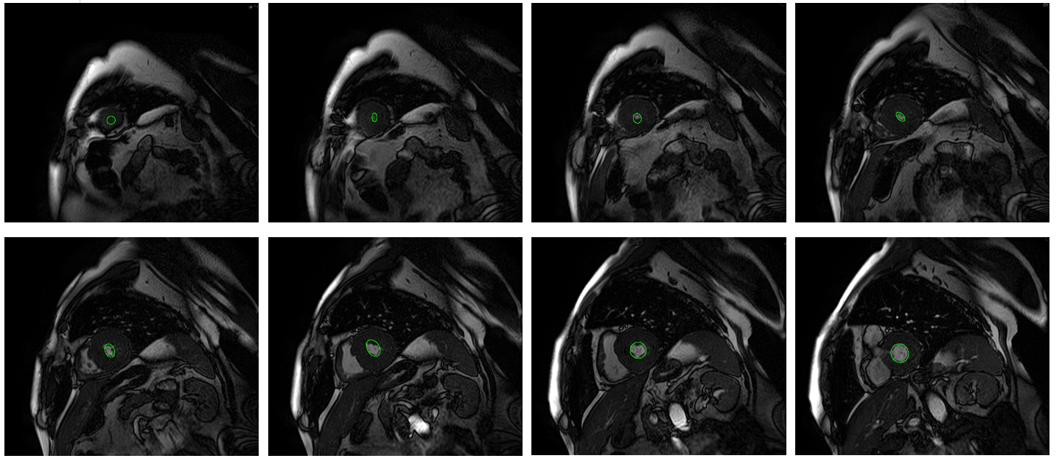}
\label{Systole2}}
\caption{(Continued) Segmented results of two different scans captured from two different scanners.}
\label{fig_res}
\end{figure*}
\section{Conclusions}
We proposed an elliptical active disc based technique for the segmentation of left ventricle in cardiac MR images. We have used multiscale template matching to localize the left ventricle irrespective of the difference in resolution of the search image or the source of the template image. Since the elliptical active disc technique requires no prior data or training it is capable of segmenting MR scans from different vendors or centers with minimal or no changes to the algorithm. The computations required for the convergence of elliptical active disc on the left ventricle are optimized using gradient descent technique and Green's theorem. We have obtained localization accuracy of left ventricle as 89.63\% and a Dice similarity coefficient of 0.873 and 0.770 on segmented diastole and systole MRI slices, respectively. 
\section*{Acknowledgments}
The authors would like to thank Prof. Chandra Sekhar Seelamantula, Department of Electrical Engineering, Indian Institute of Science, Bangalore for his guidance and support during the research. We would also like to thank the organizers of the M\&Ms challenge for providing us with a diverse collection of cardiac MR studies.

\end{document}